\documentclass[sigconf]{acmart}
\usepackage[utf8]{inputenc}
\usepackage[english]{babel}
\AtBeginDocument{%
  \providecommand\BibTeX{{%
    \normalfont B\kern-0.5em{\scshape i\kern-0.25em b}\kern-0.8em\TeX}}}

\setcopyright{iw3c2w3}
\copyrightyear{2020}
\acmYear{2020}
\acmDOI{10.1145/1122445.XXXXXXX}

\acmConference[WWW '20]{Proceedings of The Web Conference 2020}{April 20--24, 2020}{Taipei, Taiwan}
\acmBooktitle{Proceedings of The Web Conference 2020 (WWW '20), April 20--24, 2020, Taipei, Taiwan} 
\acmPrice{}
\acmISBN{978-1-4503-7023-3/20/04}

\begin{document}
\copyrightyear{2020}
\acmYear{2020} 
\acmConference[WWW '20]{Proceedings of The Web Conference 2020}{April 20--24, 2020}{Taipei, Taiwan} 
\acmBooktitle{Proceedings of The Web Conference 2020 (WWW '20), April 20--24, 2020, Taipei, Taiwan}
\acmPrice{}
\acmDOI{10.1145/3366423.3380054}
\acmISBN{978-1-4503-7023-3/20/04}

\title{VRoC: Variational Autoencoder-aided Multi-task Rumor Classifier Based on Text}



\author{Mingxi Cheng}
\affiliation{%
  \institution{University of Southern California}
}
\email{mingxic@usc.edu}

\author{Shahin Nazarian}
\affiliation{%
  \institution{University of Southern California}
}
\email{shahin.nazarian@usc.edu}


\author{Paul Bogdan}
\affiliation{%
  \institution{University of Southern California}
}
\email{pbogdan@usc.edu}



\renewcommand{\shortauthors}{Cheng, Nazarian, and Bogdan.}
\begin{abstract}
Social media became popular and percolated almost all aspects of our daily lives. While online posting proves very convenient for individual users, it also fosters fast-spreading of various rumors. The rapid and wide percolation of rumors can cause persistent adverse or detrimental impacts. Therefore, researchers invest great efforts on reducing the negative impacts of rumors. Towards this end, the rumor classification system aims to detect, track, and verify rumors in social media. Such systems typically include four components: (i) a rumor detector, (ii) a rumor tracker, (iii) a stance classifier, and (iv) a veracity classifier. In order to improve the state-of-the-art in rumor detection, tracking, and verification, we propose VRoC, a tweet-level variational autoencoder-based rumor classification
system. VRoC consists of a co-train engine that trains variational autoencoders (VAEs) and rumor classification components. The co-train engine helps the VAEs to tune their latent representations to be classifier-friendly. We also show that VRoC is able to classify unseen rumors with high levels of accuracy. For the PHEME dataset, VRoC consistently outperforms several state-of-the-art techniques, on both observed and unobserved rumors, by up to $26.9\%$, in terms of macro-F1 scores.\footnote{Code of VRoc can be found in this Github link: \url{https://github.com/cmxxx/VRoC}.} 

\end{abstract}




\maketitle
\section{Introduction}
Social media and micro-blogging have gained popularity \cite{esteban2019,Jenn2009} as tools for gathering and propagating information promptly. 
About two-thirds of Americans ($68\%$) obtain news on social media \cite{pewresearch}, while enjoying its convenient and user-friendly interfaces for learning, teaching, shopping, etc. 
Journalists use social media as convenient yet powerful tools and ordinary citizens post and propagate information via social media easily \cite{zubiaga2018detection}.
Despite the success and popularity of online media, the suitability and rapidly-spreading nature of micro-blogs fosters the emergence of various rumors \cite{cao2018automatic,allcott2019trends}. 
Individuals encountering rumors on social media may turn to other sources to evaluate, expose, or reinforce rumors \cite{lewandowsky2012misinformation,chen2015students}.
The rapid and wide spread of rumors can cause various far-reaching consequences, for example, during the 2016 U.S. presidential election, $529$ different rumors about Donald Trump and Hillary Clinton spread on social media \cite{jin2017detection} and reached millions of voters swiftly. Hence, these rumors could potentially influence the election \cite{cao2018automatic}. 
More recently, the rapid spread of rumors about 2019 novel coronavirus \cite{cohen2020new,mercey2020corona,matt2020fake} (some of which are verified to be very dangerous false claims \cite{jessica2020wuhan}, e.g., those that suggest drinking bleach cures the illness \cite{tony2020mis}) has made social media companies such as Facebook to find more effective solutions \cite{zoe2020fb}.
If not identified timely, sensational and scandalous rumors could provoke social panic during emergency events, e.g., coronavirus \cite{Julie2020}, threaten the internet credibility and trustworthiness \cite{friggeri2014rumor}, with serious implications \cite{fact2020mis}.

Social media rumors are therefore a major concern. Commercial giants, government authorities, and academic researchers heavily invest in diminishing the adverse impacts of rumors \cite{cao2018automatic}. 
The literature defines a rumor as ``an item of circulating information whose veracity status is yet to be verified at the time of posting" \cite{zubiaga2018detection}. 
On a related note, if the veracity status is confirmed to be false, the rumor can then be considered as fake news. 
Rumor handling research efforts cast four main elements: rumor detection, rumor tracking, rumor stance classification, and rumor veracity classification \cite{zubiaga2018detection}. A typical rumor classification system includes all the four elements.

\begin{figure} [t]
  \includegraphics[width=0.925\columnwidth]{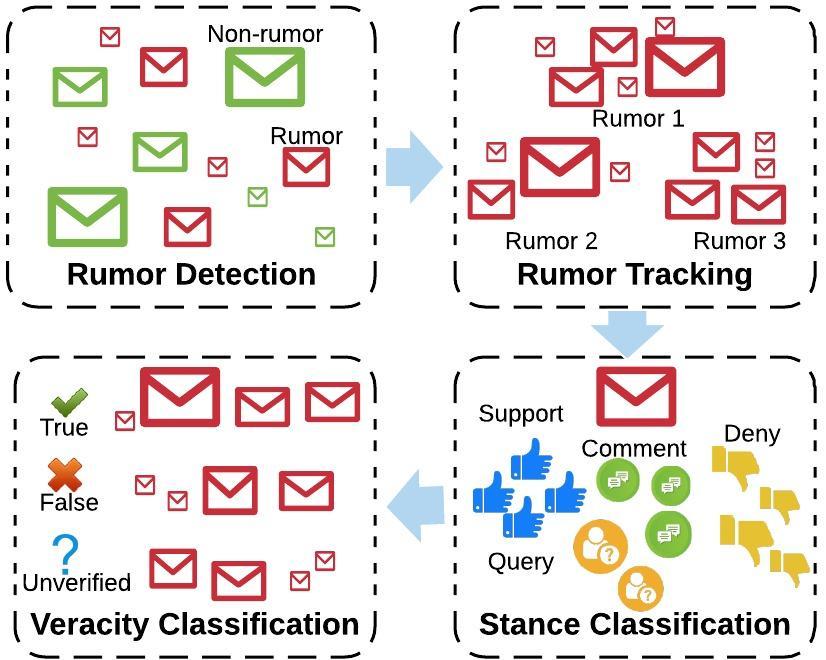}
  \caption{Rumor classification system consists of four components: rumor detection, rumor tracking, rumor stance classification, and rumor veracity classification.}
  \vskip -0.5cm
  \label{fig:rumorclassificationsystem}
\end{figure}

As shown in Fig. \ref{fig:rumorclassificationsystem}, the first step in rumor classification is rumor detection. Identifying rumors and non-rumors has been usually formulated into a binary classification problem. Among the numerous approaches, there are three major categories: hand-crafted features-based approaches, propagation-based approaches, and neural network approaches \cite{cao2018automatic}. Traditional methods mostly utilize hand-crafted features extracted from textural and/or visual contents of rumors. Having applied these features to describe the distribution of rumors, classifiers are trained to detect rumors \cite{castillo2011information,kwon2013prominent}. 
The approaches based on the structure of social network use message propagation information and evaluate the credibility of the network \cite{gupta2012evaluating}, but ignore the textual features of rumors. Social bot detection and tracking built on social network structure and user information can be utilized to detect bot-generated rumors. 
Recent deep neural network (DNN)-based methods extract and learn features automatically and achieve significantly high accuracies on rumor detection \cite{chen2018call}. Generative models and adversarial training techniques have also been used to improve the performance of rumor detectors \cite{ma2019detect}.
After a rumor is identified, all the related posts or sentences discussing this rumor should be clustered together for later processing, and other unrelated posts should be filtered out. This rumor tracking task can be formulated into a binary classification problem, which classifies posts as related or unrelated to a rumor. Unlike other popular components in rumor classification system, research in rumor tracking has been scarce. The state-of-the-art work uses tweet latent vector to overcome the limited length of a tweet \cite{hamidian2015rumor}.

Once the rumor tracking component clusters posts related to a rumor, the stance classification component labels each individual post by its orientation toward rumor's veracity. For example, a post or reply can be labeled as support, deny, comment, or query \cite{gorrell2018rumoureval}. Usually rumor stance classification can be realized as a two to four-way classification problem. 
Recurrent neural networks (RNNs) with long short-term memory (LSTM) \cite{hochreiter1997long} cells have been used to predict stance in social media conversations. The authors in \cite{kumar2019tree} proposed to use convolution units in Tree LSTMs to realize a four-way rumor stance classification. 
Variational autoencoders (VAEs) \cite{kingma2013auto} have been used to boost the performance of stance classification. The authors in \cite{sadiq2019high} utilize LSTM-based VAEs to capture the hidden meanings of rumors containing both text and images.

The final component in rumor classification is veracity classification, which determines the truth value of a rumor, i.e., a rumor can be true, false, or unverified. Some works have limited the veracity classification to binary classification, i.e., a rumor can either be true or false \cite{zhang2019reply}. 
The initiated research in this direction does not tackle the veracity of rumors directly, but rather their credibility perceptions \cite{castillo2011information}. 
Later works in this area dealing with veracity classification take advantage of temporal features, i.e., how rumors spread over time, and linguistic features.
More recently, LSTM-based DNNs are frequently used to do veracity classification \cite{kochkina2018all,kumar2019tree}. Similarly, fake news detection is often formulated into a classification problem \cite{oshikawa2018survey}. Without establishing a verified database, rumor veracity classification and fake news classification perform a similar task.

Instead of tackling each component in rumor classification system individually, multi-task classifiers are proposed to accomplish two or more functions. 
Tree LSTM models proposed in \cite{kumar2019tree} employ stance and rumor detection and propagate the useful stance signal up in the tree for followup rumor detection. 
DNNs are trained jointly in \cite{ma2018detect} to unify stance classification, rumor detection, and veracity classification tasks. 
Rumor detection and veracity classification sometimes are executed together since they can be formulated into a four-way classification problem as in \cite{ma2018detect, hamidian2015rumor}. A post can be labeled as a non-rumor, true rumor, false rumor, or unverified rumor.
The authors of \cite{kochkina2018all} proposed an LSTM-based multi-task learning approach that allows joint training of the veracity classification and auxiliary tasks, rumor detection and stance classification. 

Previous works in scientific literature have accomplished one or a few tasks in rumor classification, but none of them provides a complete high performance rumor classification system to account for all four components. In this work, we propose VRoC to realize all four tasks. The contributions of this work are as follows: 
\begin{itemize}
	\item We propose VRoC, a tweet-level text-based novel rumor classification system based on variational autoencoders. VRoC realizes all four tasks in the rumor classification system and achieves high performance compared to state-of-the-art works.
    \item We propose a co-train engine to jointly train the VAEs and rumor classification components. This engine pressurizes the VAEs to tune their latent representations to be classifier-friendly. Therefore, higher accuracies are achieved compared to other VAE-based rumor detection approach.
    \item We show that the proposed VRoC has the ability to classify previously seen or unseen rumors. Due to the generative nature of VAEs, VRoC outperforms baselines under both training policies introduced in Section \ref{exp_setting}.
\end{itemize}

The rest of this paper is organized as follows. In Section \ref{VRoC}, we introduce the proposed VRoC, including both VAE and the rumor classification system. In Section \ref{exp_setting}, we describe the dataset and baselines we used in our experiments. In Section \ref{exp_results}, we show our experimental results. Section \ref{Conclusion} concludes the paper.


\begin{figure*}
  \includegraphics[width=\textwidth]{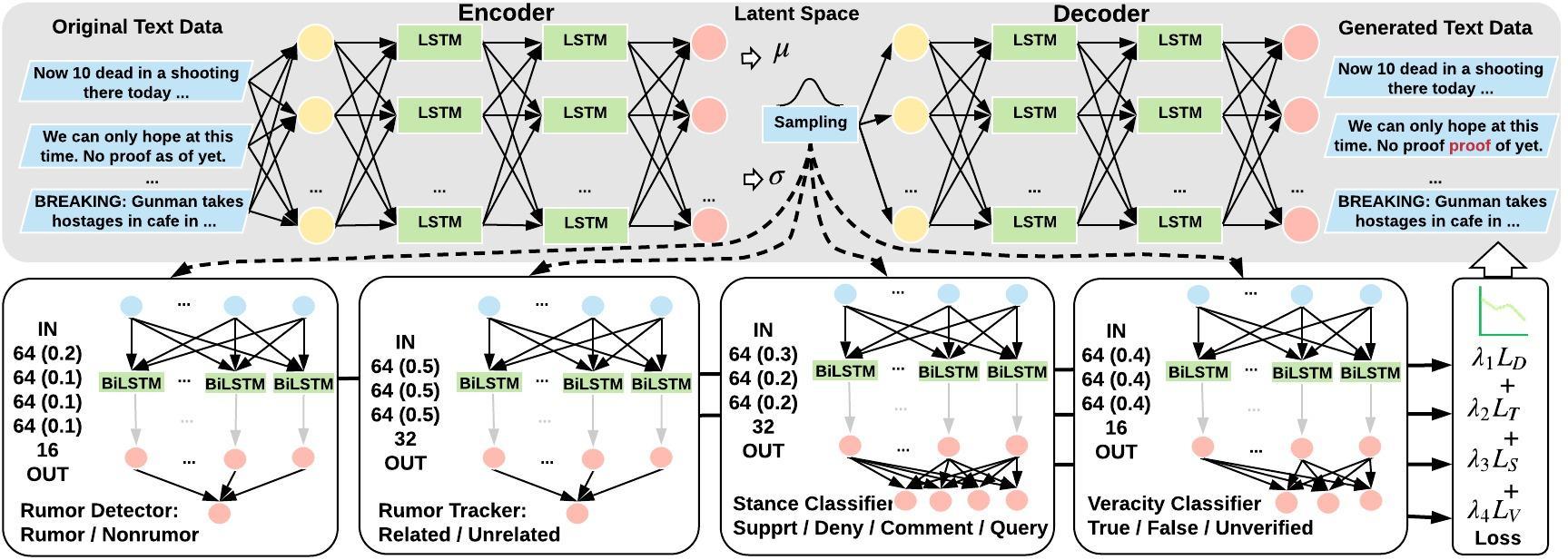}
  \caption{VRoC: The proposed VAE-aided multi-task rumor classification system. The top half illustrates the VAE structure, and the bottom half shows the four components in the rumor classification system. IN and OUT represent input layer and output layer, respectively. Numbers in parenthesis indicate the dropout rates. Note that the generated text could be different from the original text, if the VAE is not perfect. }
  \vskip -0.3cm
  \label{fig:VRoC}
\end{figure*}

\section{VR\lowercase{o}C Framework} \label{VRoC}
In this section we first present the problem statement and then describe the details of our VRoC framework.
Fig. \ref{fig:VRoC} illustrates our VRoC, a VAE-aided multi-task rumor classifier that consists of rumor detector, rumor tracker, stance classifier, and veracity classifier.
VAE in this work is an LSTM-based variational autoencoder model to extract latent representations of tweet-level text. 
For each rumor classification component, a VAE is jointly trained to extract meaningful latent representations that not only is information rich, but also is friendly and suitable for each component. 

\subsection{Problem Statement}
Rumor classification consists of four components: rumor detection (D), rumor tracking (T), stance (S) classification, and veracity (V) classification, each of which can be formulated into a classification problem. 
Given a tweet-level text $x$, VRoC can provide four outputs $\{y_D,y_T,y_S,y_V\}$, where $y_D\in\{Rumor,Nonrumor\}$, $y_T\in\{Related,Unrelated\}$, $y_S\in\{Support, Deny, Comment, Query\}$, $y_V\in\{True, False, Unverified\}$. 
Four components are realized independently in this work, but they can also be implemented as one general classifier that jointly produces four types of outputs, or two to three classifiers that each realizes one or more tasks.

\subsection{LSTM-based Variational Autoencoder}
The VAE model in this work consists of an encoder and a decoder, both of which are LSTM networks because of the sequential nature of language. 
To extract latent representation from tweet-level data, 
in this work, we take advantage of VAEs and utilize the encoder-extracted latent representations to compress and represent information in textural data. 
The decoder decodes the latent representations generated by the encoder into texts and ensures the latent representations are accurate and meaningful.
Rumor classifier components are co-trained with VAEs to help the encoders tune their outputs to be more classifier-friendly.

Let us consider a set of tweets $X=\{x_1,x_2,...,x_N\}$, each of which is generated by some random process $p_{\theta}(x|z)$ with an unobserved variable $z$. This $z$ is generated by a prior distribution $p_{\theta}(z)$ that is hidden from us. 
In order to classify or utilize these tweets, we have to infer the marginal likelihood $p_{\theta}(x)$, however $\theta$ is unfortunately also unknown to us.
In VAEs, $z$ is drawn from a normal distribution, and we attempt to find a function $p_{\theta}(x|z)$ that can map $z$ to $x$ by optimizing $\theta$, such that $x$ looks like what we have in data $X$.
From coding theory, $z$ is a latent representation and we can use a recognition function $q_{\phi}(z|x)$ as the encoder \cite{kingma2013auto}. $q_{\phi}(z|x)$ takes a value $x$ and provides the distribution over $z$ that is likely to produce $x$. $q_{\phi}(z|x)$ is also an approximation to the intractable true posterior $p_{\theta}(z|x)$. 
Kullback-Leibler (KL) divergence measures how close the $p_{\theta}(z|x)$ and $q_{\phi}(z|x)$ are:
\begin{equation}
    D_{KL}[q_{\phi}(z|x)||p_{\theta}(z|x)] = E_{z\sim q_{\phi}(z|x)}[log q_{\phi}(z|x)-log p_{\theta}(z|x)].
\end{equation}
After applying Bayesian theorem, the above equation reads:
\begin{equation}
    log p_{\theta}(x) = 
    D_{KL}[q_{\phi}(z|x)||p_{\theta}(z|x)] + L(\theta, \phi;x),
\end{equation}
\begin{equation} \label{eq:lowerbound}
    L(\theta, \phi;x) = E_{z\sim q_{\phi}(z|x)}[-log q_{\phi}(z|x)+log p_{\theta}(x|z)+log p_{\theta}(z)].
\end{equation}
We can see the autoencoder from Eq. \ref{eq:lowerbound} already: $q_{\phi}$ encodes $x$ into $z$ and $p_{\theta}$ decodes $z$ back to $x$. 
Since KL divergence is non-negative, $L(\theta, \phi;x)$ is called the evidence lower bound (ELBO) of $log p_{\theta}(x)$:
\begin{equation}
    log p_{\theta}(x) \geq  L(\theta, \phi;x).
\end{equation}
In VAEs, $q_{\phi}(z|x)$ is a Gaussian: $N(z;\mu,\sigma^2I)$, where $\mu$ and $\sigma$ are outputs of the encoder.
The reparameterization trick \cite{kingma2013auto} is used to express $z\sim q_{\phi}(z|x)$ as a random variable $z = g_{\phi}(\epsilon,x) = \mu + \sigma \cdot \epsilon$, where the auxiliary variable $\epsilon$ is drawn from a standard normal distribution. 
A Monte Carlo estimation is then formed for ELBO as follows:
\begin{equation}
    L^{MC}(\theta, \phi;x) = 
   \frac{1}{N}\sum_{n=1}^N -log  q_{\phi}(z_n|x)+log p_{\theta}(x|z_n)+log p_{\theta}(z_n),
\end{equation}
where $z_n = g_{\phi}(\epsilon_n,x)$, $\epsilon_n \sim N(0,I)$.

\subsubsection{Encoder}
The encoder $q_{\phi}(z|x)$ is realized by an RNN with LSTM cells. The input to the encoder is a sequence of words $x = [w_1,w_2,...,w_T]$, e.g., a tweet with length $T$. 
The hidden state $h_t$, $t\in[1,T]$ in LSTM is updated as follows: 
\begin{equation}
    h_t = o_t*tanh(C_t),
\end{equation}
\begin{equation}
    o_t = \sigma(W_o\cdot[h_{t-1},w_t]+b_o),
\end{equation}
\begin{equation}
    C_t = f_t*C_{t-1}+i_t*\bar{C_t},
\end{equation}
\begin{equation}
    f_t = \sigma(W_f\cdot[h_{t-1},w_t]+b_f),
\end{equation}
\begin{equation}
    i_t = \sigma(W_i\cdot[h_{t-1},w_t]+b_i),
\end{equation}
\begin{equation}
    \bar{C_t} = tanh(W_C[h_{t-1},w_t]+b_C).
\end{equation}
$\sigma$ is a sigmoid function. $o_t$, $f_t$, $i_t$ are output, forget, and input gate, respectively. $C_t$, $\bar{C_t}$ are new state and candidate cell state, respectively. $W_o$, $W_f$, $W_i$, $W_C$, $b_o$, $b_f$, $b_i$, $b_C$ are parameter metrics. 
The outputs of the encoder are divided into $\mu$ and $\sigma$.

\subsubsection{Decoder}
The decoder $p_{\phi}(x|z)$ is realized by an RNN with LSTM cells which has a matching architecture as that of the encoder. It takes a $z$ as input and outputs the probabilities of all words. The probabilities are then sampled and decoded into a sequence of words.

\subsubsection{Training}
In this work, we propose a \textit{co-train engine} for VAEs. Each component in rumor classifier is trained jointly with a VAE, i.e., there are four sets of VAEs and components. Each set is trained individually. 
To co-train each set, we modify the loss function of VAE by adding a classification penalty, which is the loss of the rumor classification component. By backpropagating the loss of each component to the corresponding VAE, the VAE learns to tune its parameters to provide classifier-friendly latent representations. In addition, this operation introduces randomness into the training process of VAEs, hence the robustness and generalization ability of VRoC are improved.
The loss function of VRoC then reads:
\begin{equation}
    L_{VRoC} = L^{MC}(\theta, \phi;x) + \lambda_1L_{D} + \lambda_2L_{T} + \lambda_3L_{S} + \lambda_4L_{V},
\end{equation}
where $\lambda_1$, $\lambda_2$, $\lambda_3$, $\lambda_4$ are balancing parameters for each rumor classification component. $L_{D}$, $L_{T}$, $L_{S}$, $L_{V}$ are loss functions of rumor detector, rumor tracker, stance classifier, and veracity classifier, respectively.
Training VAEs and the components jointly improves the performance of the rumor classifier, compared to the VAE-based rumor classifiers without our co-train engine. 
We confirm the improvement by conducting a set of comparison experiments between VRoC and VAE-LSTM in Section \ref{exp_results}.

\subsection{Rumor Classifier}
In this section, we introduce all four components in VRoC's rumor classification system and their loss functions. The loss functions are consistent with free-energy minimization principle \cite{friston2009free}.
\subsubsection{Rumor Detector}
Given a set of social media posts $X=\{x_1,x_2,...,x_N\}$, rumor detector determines which ones are rumors and which ones are not. The classified posts can then be tracked and verified in later steps. In this work, rumor detection task is formulated as a binary classification problem. Rumor detector $D$ is realized by an RNN with Bidirectional-LSTM (BiLSTM) cells. It takes a $z$ as input and provides a probability $y_D$ of the corresponding post $x$ being a rumor. The loss function $L_D$ is defined as follows: 
\begin{equation}
    L_D = -E_{y\sim Y_D}[ylog(y_D)+(1-y)log(1-y_D)],
\end{equation}
where $Y_D \in \{Rumor, Nonrumor\}$ is a set of ground truth labels.

\subsubsection{Rumor Tracker}
Rumor tracker $T$ is activated once a rumor is identified. It takes a set of posts $X$ as input, and determines whether each post is related or unrelated to the given rumor. Rumor tracking task is formulated into a classification problem in this work, and it is fulfilled by an RNN with BiLSTM cells. Given a $z$, $T$ generates a probability $y_T$ indicating whether the corresponding post is related to the identified rumor. The loss function $L_T$ reads:
\begin{equation}
    L_T = -E_{y\sim Y_T}[ylog(y_T)+(1-y)log(1-y_T)],
\end{equation}
where $Y_T \in \{Related, Unrelated\}$.

\subsubsection{Stance Classifier}
Given a collection of rumors $R =\{r_1,r_2,...,r_N\}$, where each rumor $r_n$ ($n\in[1,N]$) consists of a set of posts discussing it, the stance classifier $S$ determines whether each post is supporting, denying, commenting, or querying the related rumor $r_n$. In VRoC, we utilize an RNN with BiLSTM cells to perform a four-way rumor stance classification. The loss function $L_S$ is defined as follows:
\begin{equation}
    L_S = -E_{y\sim Y_S}[\sum ylog(y_S)],
\end{equation}
where $Y_S \in \{Support, Deny, Comment, Query\}$.

\subsubsection{Veracity Classifier}
Once rumors are identified, their truth values are determined by rumor veracity classifier $V$. Instead of being true or false, some rumors could in reality remain unverified for a period of time. Hence, in this work, we provide a three-way veracity classification using an RNN with BiLSTM cells. The loss function $L_V$ is defined as follows:
\begin{equation}
    L_V = -E_{y\sim Y_V}[\sum ylog(y_V)],
\end{equation}
where $Y_V \in \{True, False, Unverified\}$. This veracity classifier can also be used for fake news detection since it performs a similar task.

\section{Experimental Settings} \label{exp_setting}
In this section, we first describe the datasets and evaluation metrics, then introduce the baselines from state-of-the-art works.

\subsection{Datasets and Evaluation Metrics}
We evaluate our VRoC on PHEME dataset \cite{kochkina2018all}. PHEME has two versions. PHEME5 contains $5792$ tweets related to five news, $1972$ of them are rumors and $3820$ of them are non-rumors. PHEME9 is extended from PHEME5 and contains veracity labels. 
RumourEval dataset \cite{gorrell2018rumoureval} is derived from PHEME, and its stance labels are used for rumor stance classification task. 
We use PHEME5 for rumor detection and tracking task, PHEME5 with stance labels from RumourEval for rumor stance classification task, and PHEME5 with veracity labels from PHEME9 for the rumor veracity classification task.
Due to the imbalance of classes in dataset, accuracy alone as evaluation metric is not sufficient. Hence, we use precision, recall, and macro-F1 scores \cite{van2013macro,li2010data} together with accuracy as evaluation metrics. 
Since baselines are trained under different principles, we carry out two types of trainings to guarantee fairness. 
To compare with baselines that are trained under the leave-one-out (L) principle, i.e., train on four news and test on another news, we train our models under the same principle. L principle evaluates the ability of generalization and constructs a test environment close to real world scenarios.
To compare with baselines that do not use this principle, we hold out $10\%$ of the data for model tuning.

\subsection{Models}
In \textbf{VRoC}, we co-train each set of VAE and classification component after pre-training the VAEs. 
To show the effectiveness of our designed co-train engine, we developed a baseline \textbf{VAE-LSTM} that trains VAE first, then uses its latent representations as input to train the LSTM models. VAE-LSTM's architecture is the same as VRoC, but without the proposed co-train engine.
The VAEs used in VRoC and VAE-LSTM are pre-trained with the same procedure. The encoder architecture in VAE is EM-LSTM32-LSTM32-D32, where EM, LSTM32, D32 represent an embedding layer, an LSTM layer with $32$ neurons, and a dense layer with $32$ neurons, respectively. The decoder's architecture is IN-LSTM32-LSTM32-Dvs, where IN and Dvs represent input layer and dense layer with vocabulary size neurons. 
We chose our VAE architecture by an expert-guided random search (RS) under the consideration of data size. Compared to other computationally expensive neural architecture search methods, such as evolutionary search and reinforcement learning-based approaches, RS is confirmed to achieve competitive performance \cite{elsken2018neural, real2019regularized, liu2017hierarchical}. Early stopping strategy is used in training. 
Other state-of-the-art baselines used for comparison are described as follows.

\subsubsection{Rumor Detection}
\textbf{CRF} \cite{zubiaga2017exploiting} is a content and social feature-based rumor detector that based on linear-chain conditional random fields learns the dynamics of information during breaking news.
\textbf{GAN-GRU}, \textbf{GAN-BOW}, and \textbf{GAN-CNN} \citep{ma2019detect} are generative adversarial network-based rumor detectors. A generator is designed to create uncertainty and the complicated sequences force the discriminator to learn stronger rumor representations. 
\textbf{DataAUG} \cite{han2019data} is a contextual embedding model with data augmentation. It exploits the semantic relations between labeled and unlabeled data. 
\textbf{DMRF} \cite{nguyen2019fake} formulates rumor detection task as an inference problem in a Markov Random Field (MRF). It unfolds the mean-field algorithm into neural network and builds a deep MRF model.
\textbf{DTSL} \cite{dong2019deep} is a deep semi-supervised learning model containing three CNNs. 

\subsubsection{Rumor Tracking}
As we mentioned before, rumor tracking receives few attention in scientific literature. Hence, we train two baselines to compare with our proposed VRoC. 
\textbf{CNN} is a convolutional neural network-based baseline. \textbf{LSTM} is an RNN with LSTM cells. We input all five news in PHEME to all models, and perform a five-way classification, i.e., determine the input post is related to which one in five news. Five news are: Sydney siege (S), Germanwings (G), Ferguson (F), Charlie Hebdo (C), and Ottawa shooting (O). L principle is not applicable under this setup.

\subsubsection{Stance Classification}
\textbf{LinearCRF} and \textbf{TreeCRF} are two different models proposed in \cite{zubiaga2016stance} for capturing the sequential structure of conversational threads. They analyse tweets by mining the context from conversational threads. 
The authors in \cite{ma2018detect} proposed a unified multi-task (rumor detection and stance classification) model based on multi-layer RNNs, where a shared layer and a task-specific layer are employed to accommodate different types of representations of the tasks and their corresponding parameters. \textbf{MT-US} and \textbf{MT-ES} are multi-task models with the uniform shared-layer and enhanced shared-layer architecture.

\subsubsection{Veracity Classification}
\textbf{MT-UA} \cite{li2019rumor} is a multi-task rumor detector that utilizes the user credibility information and attention mechanism. 
In \cite{kochkina2018all}, a joint multi-task deep learning model with hard parameters sharing is presented. It outperforms the sequential models by combining multiple tasks together. We choose two best performing models from \cite{kochkina2018all}: \textbf{MTL2} that combines veracity and stance classification, and \textbf{MTL3} that combines detection, stance, and veracity classification. 
\textbf{TreeLSTM} \cite{kumar2019tree} is a tree LSTM model that uses convolution and max-pooling unit.

\section{Experimental Results} \label{exp_results}
The comparison between VRoC and baselines in all four rumor classification tasks are presented in Section \ref{exp:d} to \ref{exp:v}. In all tables, * indicates the best results from the work that proposed the corresponding model. We finally describe the comparison results of VRoC and VAE-LSTM in Section \ref{exp:vaelstm}.

\subsection{Rumor Detection} \label{exp:d}
Comparison results between VRoC and baselines on the rumor detection task are shown in Table \ref{table:detection}. Compared to baselines, VRoC achieves significantly higher accuracy levels and macro-F1 scores, and VAE-LSTM stands as the second best. VRoC outperforms CRF and GAN-GRU by $26.9\%$ and $9.5\%$ in terms of macro-F1. Under L principle, on average, VRoC and VAE-LSTM outperforms baselines by $13.2\%$ and $14.9\%$ in terms of macro-F1. VAE's compact latent representations contribute to these results the most. Compared to VAE-LSTM, the proposed co-train engine boosts the performance of VRoC one step further.

\begin{table}[h]
\centering 
\caption{Comparison between VRoC and baselines on the rumor detection task. }\label{table:detection}
	\resizebox{0.9\columnwidth}{!}{
		\begin{tabular}{l c c c c c }
			\toprule
		         & Accuracy & Precision & Recall & Macro-F1 \\\hline
			CRF* & - & $0.667$ & $0.556$ & $0.607$ \\
			GAN-BOW* & $0.781$ & $0.782$ & $0.781$ & $0.781$ \\
			GAN-CNN* & $0.736$ & $0.738$ & $0.736$ & $0.736$ \\
			GAN-GRU* & $0.688$ & $0.689$ & $0.688$ & $0.687$ \\
			VAE-LSTM & $0.833$ & $0.834$ & $0.834$ & $0.833$ \\
			VRoC & $\textbf{0.876}$ & $\textbf{0.877}$ & $\textbf{0.876}$ & $\textbf{0.876}$\\\hline
			DataAUG* (L) & $0.707$ & $0.580$ & $0.497$ & $0.535$\\
			DMFN* (L)& $0.703$ & $0.667$ & $0.670$ & $0.657$\\
			DTSL* (L) & - & $0.560$ & $\textbf{0.794}$ & $0.615$ \\
			VAE-LSTM (L) & $0.736$ & $0.746$ & $0.736$ & $0.735$ \\
			VRoC (L) & $\textbf{0.752}$ & $\textbf{0.755}$ & $0.752$ & $\textbf{0.752}$\\
            \bottomrule
		\end{tabular}
    }
\end{table}

\subsection{Rumor Tracking} \label{exp:t}
Comparison results between VRoC and baselines on the rumor tracking task are shown in Table \ref{table:tracking}. 
VRoC achieves the highest macro-F1, but it does not outperform baselines by a large percentage. In rumor tracking, raw data might be a preferable data source since they contain keywords and hashtags that can be used to directly track the topic. For long posts, rumor tracking can be effortlessly accomplished by retrieving the hashtags. Compared to models that use raw data, VRoC has advantages when dealing with imbalanced and unseen data since it can extract compact information from a few posts and generalize to a broader range of data. 
\begin{table}[h]
\centering 
\caption{Comparison between VRoC and baselines on the rumor tracking task. }\label{table:tracking}
	\resizebox{\columnwidth}{!}{
		\begin{tabular}{l c c c c c c c c }
			\toprule
			& & & S & G & F & C & O \\
			& Accuracy & Macro-F1 & F1 & F1 & F1 & F1 & F1 \\\hline
			CNN & $0.570$ & $0.574$ & $0.589$ & $\textbf{0.534}$ & $0.777$ & $0.571$ & $0.400$ \\
			LSTM & $0.585$ & $0.585$ & $0.607$ & $0.352$ & $\textbf{0.804}$ & $\textbf{0.711}$ & $0.453$\\
			VAE-LSTM & $0.609$ & $0.612$ & $\textbf{0.666}$ & $0.515$ & $0.641$ & $0.694$ & $0.545$ \\
			VRoC & $\textbf{0.644}$ & $\textbf{0.632}$ & $0.611$ & $0.520$ & $0.640$ & $0.685$ & $\textbf{0.703}$ \\
            \bottomrule
		\end{tabular}
    }
\end{table}

\subsection{Stance Classification} \label{exp:s}
Comparison results between VRoC and baselines on the rumor stance classification task are shown in Table \ref{table:stance}. Stance classification is the hardest component in rumor classification. The reason is two-fold: four-way classification problems are naturally more difficult than binary classification problems, and imbalanced data exaggerate the difficulty. 
In addition, the stance classification is not as easy as the tracking task. Stance classifier has to extract detailed patterns from a sentence and consider the whole sentence. 
In the tracking task, posts can be classified together by filtering out the obvious keywords, but stance is related to the semantic meaning of the whole sentence and hence is more complicated. 
Baselines in the comparison all suffer from the extremely low F1 score on Deny class, which is caused by the small size of the Deny instances in the dataset. Feature extraction from raw data results in severely imbalanced performance among different classes. VRoC's and VAE-LSTM's classifiers are trained under latent representations. Although data imbalance affects the performance of VRoC and VAE-LSTM, the impact is not as drastic as in other baselines. Compared to state-of-the-art baselines that concentrate on stance classification, VRoC's stance classification component provides the highest macro-F1 scores under both training principles and VAE-LSTM follows as the second best. 

\begin{table}[h]
\centering 
\caption{Comparison between VRoC and baselines on the rumor stance classification task. }\label{table:stance}
	\resizebox{\columnwidth}{!}{
		\begin{tabular}{l c c c c c c}
			\toprule
			& & & Support & Deny & Comment & Query\\
			& Accuracy & Macro-F1 & F1 & F1 & F1 & F1 \\\hline
			MT-US* & - & $0.400$ & $0.355$ & $0.116$ & $\textbf{0.776}$ & $0.337$  \\
			MT-ES* & - & $0.430$ & $0.314$ & $0.158$ & $0.739$ & $\textbf{0.531}$  \\
			VAE-LSTM & $0.464$ & $0.461$ & $0.447$ & $0.395$ & $0.588$ & $0.416$ \\
			VRoC & $\textbf{0.533}$ & $\textbf{0.522}$ & $\textbf{0.452}$ & $\textbf{0.415}$ & $0.712$ & $0.511$\\\hline
			TreeCRF* (L) & - & $0.440$ & $0.462$ & $0.088$ & $\textbf{0.773}$ & $0.435$ \\
			LinearCRF* (L) & - & $0.433$ & $0.454$ & $0.105$ & $0.767$ & $\textbf{0.495}$  \\
			VAE-LSTM (L) & $0.467$ & $0.459$ & $\textbf{0.463}$ & $0.423$ & $0.567$ & $0.384$ \\
			VRoC (L) & $\textbf{0.480}$ & $\textbf{0.473}$ & $0.452$ & $\textbf{0.429}$ & $0.596$ & $0.416$\\
            \bottomrule
		\end{tabular}
    }
\end{table}

\subsection{Veracity Classification} \label{exp:v}
Comparison results between VRoC and baselines on the rumor veracity  classification task are shown in Table \ref{table:veracity}. VRoC and VAE-LSTM achieve the highest macro-F1 and accuracy compared to baselines. 
On average, VRoC outperforms MT-UA and other baselines under L principle by $24.9\%$ and $11.9\%$ in terms of macro-F1, respectively.
Rumor veracity classification under L principle is particularly difficult since there is no previously established verified news database. 
An unseen news on a previously unobserved event has to be classified without any knowledge related to this event. For example, you observed some news on A event, and you need to verify whether a news from an unrelated B event is true or not. 
Without a verified news database, the abstracted textural patterns of sentences are utilized to classify unobserved news. Latent representations extracted by VAEs are hence very helpful to generalize in veracity classification. 
The outperformance of VRoC and VAE-LSTM over baselines under L principle demonstrates the outstanding generalization ability of VAEs. In addition, VRoC beats VAE-LSTM in terms of both accuracy and macro-F1 in all cases. These results further demonstrate the power of the proposed co-train engine.

\begin{table}[h]
\centering
\caption{Comparison between VRoC and baselines on the rumor veracity classification task. Lc represents that the news related to Charlie Hebdo is left out while trained under L principle. }\label{table:veracity}
	\resizebox{\columnwidth}{!}{
		\begin{tabular}{l c c c c c}
			\toprule
			& & & True & False & Unverified \\
			& Accuracy & Macro-F1 & F1 & F1 & F1  \\\hline
			MT-UA* & $0.483$ & $0.418$ & - & - & -\\
			VAE-LSTM & $0.628$ & $0.627$ & $0.691$ & $0.576$ & $0.615$  \\
			VRoC & $\textbf{0.667}$ & $\textbf{0.667}$ & $\textbf{0.745}$ & $\textbf{0.632}$ & $\textbf{0.624}$\\\hline
			MTL2* (Lc) & $0.441$ & $0.376$ & - & - & -   \\
			MTL3* (Lc) & $0.492$ & $0.396$ & $\textbf{0.681}$ & $0.232$ & $0.351$  \\
			VAE-LSTM (Lc) & $0.507$ & $0.503$ & $0.545$ & $\textbf{0.449}$ & $\textbf{0.515}$  \\
			VRoC (Lc)& $\textbf{0.531}$ & $\textbf{0.513}$ & $0.564$ & $0.434$ & $0.480$\\\hline
			TreeLSTM* (L) & $0.500$ & $0.379$ & $0.396$ & $\textbf{0.563}$ & $0.506$ \\
			VAE-LSTM (L) & $0.494$ & $0.475$ & $0.429$ & $0.472$ & $\textbf{0.523}$  \\
			VRoC (L) & $\textbf{0.521}$ & $\textbf{0.484}$ & $\textbf{0.480}$ & $0.504$ & $0.465$\\
			
            \bottomrule
		\end{tabular}
    }
\end{table}

\subsection{VRoC and VAE-LSTM} \label{exp:vaelstm}
As shown in Tables \ref{table:detection}-\ref{table:veracity}, VRoC outperforms all the baselines in terms of macro-F1 and accuracy in all four rumor classification tasks, while VAE-LSTM stands as the second best. On average, in all four tasks, VRoC and VAE-LSTM surpass the baselines by $10.94\%$ and $7.64\%$ in terms of macro-F1 scores. The ability of VAE-based rumor classifier is confirmed by these results. The advantage of latent representations over raw tweet data is demonstrated as well. 
VRoC achieves higher performance than VAE-LSTM because of the designed co-train engine. 
VRoC's latent representations are more suitable and friendly to the rumor classification components. Furthermore, the co-train engine introduces randomness into the training process of VRoC, hence the robustness and generalization abilities of VRoC are improved. Dimensionality reduction \cite{an2015variational} is also realized by the VAEs to further aid the generalization. Semantically and syntactically related samples are placed near each other in latent space. Although future news are unobserved, they may contain similar semantic and/or syntactic features to those observed news. Thus VRoC could generalize and place the new latent representations close to the old ones and classify them without the need of retrain. VRoC and VAE-LSTM are efficient since all four tasks can be performed in parallel. Assume the serial runtime is $T_s$, the parallel runtime is $T_p=\frac{T_s}{p}$ (if all four tasks are parallelized and $p=4$ is number of processors used in parallel), then the efficiency $E=\frac{Speedup}{p}=\frac{T_s/T_p}{p}=1$.

\section{Conclusions} \label{Conclusion}
Various rumors on social media during emergency events threaten the internet credibility and provoke social panic, and may lead to long-term negative consequences. Recent rumors on the 2019 novel coronavirus stand as a shocking example.
To mitigate such issues provoked by rumors, we propose VRoC, a variational autoencoder-aided rumor classification system consisting of four components: rumor detection, rumor tracking, stance classification, and veracity classification. 
The novel architecture of VRoC, including its suitable classification techniques in the tasks associate with rumor handling and the designed co-train engine contribute to the high performance and generalization abilities of VRoC. 
Facing previously observed or unobserved rumors, VRoC outperforms state-of-the-art works by $10.94\%$ in terms of macro-F1 scores on average. 
VRoC is efficient not only in the sense of parallel computing, but also in terms of speed of rumor detection. The high accuracy shortens the time of detection and hence helps with reducing the chance of rumor resurfacing. 
As part of our future work, we would like to investigate the cooperation of four components to improve the overall performance. We will also extend our work by including visual contents such as images, and explore the influence between text and visual features in rumors. 

\begin{acks}
\vskip 0.5cm
The authors gratefully acknowledge the support by the National Science Foundation under the Career Award CPS/CNS-1453860, CCF-1837131, MCB-1936775, CNS-1932620, and the DARPA Young Faculty Award, under grant number N66001-17-1-4044. The views, opinions, and/or findings contained in this article are those of the authors and should not be interpreted as representing the official views or policies, either expressed or implied by the Defense Advanced Research Projects Agency, the Department of Defense or the National Science Foundation.
\vskip 0.5cm
\end{acks}






\bibliographystyle{ACM-Reference-Format}
\bibliography{bibli}


\begin{thebibliography}{46}


\ifx \showCODEN    \undefined \def \showCODEN     #1{\unskip}     \fi
\ifx \showDOI      \undefined \def \showDOI       #1{#1}\fi
\ifx \showISBNx    \undefined \def \showISBNx     #1{\unskip}     \fi
\ifx \showISBNxiii \undefined \def \showISBNxiii  #1{\unskip}     \fi
\ifx \showISSN     \undefined \def \showISSN      #1{\unskip}     \fi
\ifx \showLCCN     \undefined \def \showLCCN      #1{\unskip}     \fi
\ifx \shownote     \undefined \def \shownote      #1{#1}          \fi
\ifx \showarticletitle \undefined \def \showarticletitle #1{#1}   \fi
\ifx \showURL      \undefined \def \showURL       {\relax}        \fi
\providecommand\bibfield[2]{#2}
\providecommand\bibinfo[2]{#2}
\providecommand\natexlab[1]{#1}
\providecommand\showeprint[2][]{arXiv:#2}

\bibitem[\protect\citeauthoryear{Allcott, Gentzkow, and Yu}{Allcott
  et~al\mbox{.}}{2019}]%
        {allcott2019trends}
\bibfield{author}{\bibinfo{person}{Hunt Allcott}, \bibinfo{person}{Matthew
  Gentzkow}, {and} \bibinfo{person}{Chuan Yu}.}
  \bibinfo{year}{2019}\natexlab{}.
\newblock \showarticletitle{Trends in the diffusion of misinformation on social
  media}.
\newblock \bibinfo{journal}{\emph{Research \& Politics}} \bibinfo{volume}{6},
  \bibinfo{number}{2} (\bibinfo{year}{2019}),
  \bibinfo{pages}{2053168019848554}.
\newblock


\bibitem[\protect\citeauthoryear{An and Cho}{An and Cho}{2015}]%
        {an2015variational}
\bibfield{author}{\bibinfo{person}{Jinwon An} {and} \bibinfo{person}{Sungzoon
  Cho}.} \bibinfo{year}{2015}\natexlab{}.
\newblock \showarticletitle{Variational autoencoder based anomaly detection
  using reconstruction probability}.
\newblock \bibinfo{journal}{\emph{Special Lecture on IE}} \bibinfo{volume}{2},
  \bibinfo{number}{1} (\bibinfo{year}{2015}).
\newblock


\bibitem[\protect\citeauthoryear{Cao, Guo, Li, Jin, Guo, and Li}{Cao
  et~al\mbox{.}}{2018}]%
        {cao2018automatic}
\bibfield{author}{\bibinfo{person}{Juan Cao}, \bibinfo{person}{Junbo Guo},
  \bibinfo{person}{Xirong Li}, \bibinfo{person}{Zhiwei Jin},
  \bibinfo{person}{Han Guo}, {and} \bibinfo{person}{Jintao Li}.}
  \bibinfo{year}{2018}\natexlab{}.
\newblock \showarticletitle{Automatic rumor detection on microblogs: A survey}.
\newblock \bibinfo{journal}{\emph{arXiv preprint arXiv:1807.03505}}
  (\bibinfo{year}{2018}).
\newblock


\bibitem[\protect\citeauthoryear{Castillo, Mendoza, and Poblete}{Castillo
  et~al\mbox{.}}{2011}]%
        {castillo2011information}
\bibfield{author}{\bibinfo{person}{Carlos Castillo}, \bibinfo{person}{Marcelo
  Mendoza}, {and} \bibinfo{person}{Barbara Poblete}.}
  \bibinfo{year}{2011}\natexlab{}.
\newblock \showarticletitle{Information credibility on twitter}. In
  \bibinfo{booktitle}{\emph{Proceedings of the 20th international conference on
  World wide web}}. ACM, \bibinfo{pages}{675--684}.
\newblock


\bibitem[\protect\citeauthoryear{Chen, Li, Yin, and Zhang}{Chen
  et~al\mbox{.}}{2018}]%
        {chen2018call}
\bibfield{author}{\bibinfo{person}{Tong Chen}, \bibinfo{person}{Xue Li},
  \bibinfo{person}{Hongzhi Yin}, {and} \bibinfo{person}{Jun Zhang}.}
  \bibinfo{year}{2018}\natexlab{}.
\newblock \showarticletitle{Call attention to rumors: Deep attention based
  recurrent neural networks for early rumor detection}. In
  \bibinfo{booktitle}{\emph{Pacific-Asia Conference on Knowledge Discovery and
  Data Mining}}. Springer, \bibinfo{pages}{40--52}.
\newblock


\bibitem[\protect\citeauthoryear{Chen, Sin, Theng, and Lee}{Chen
  et~al\mbox{.}}{2015}]%
        {chen2015students}
\bibfield{author}{\bibinfo{person}{Xinran Chen},
  \bibinfo{person}{Sei-Ching~Joanna Sin}, \bibinfo{person}{Yin-Leng Theng},
  {and} \bibinfo{person}{Chei~Sian Lee}.} \bibinfo{year}{2015}\natexlab{}.
\newblock \showarticletitle{Why students share misinformation on social media:
  Motivation, gender, and study-level differences}.
\newblock \bibinfo{journal}{\emph{The Journal of Academic Librarianship}}
  \bibinfo{volume}{41}, \bibinfo{number}{5} (\bibinfo{year}{2015}),
  \bibinfo{pages}{583--592}.
\newblock


\bibitem[\protect\citeauthoryear{Cohen and Normile}{Cohen and Normile}{2020}]%
        {cohen2020new}
\bibfield{author}{\bibinfo{person}{Jon Cohen} {and} \bibinfo{person}{Dennis
  Normile}.} \bibinfo{year}{2020}\natexlab{}.
\newblock \bibinfo{title}{New SARS-like virus in China triggers alarm}.
\newblock
\newblock


\bibitem[\protect\citeauthoryear{Dong, Victor, Chowdhury, and Qian}{Dong
  et~al\mbox{.}}{2019}]%
        {dong2019deep}
\bibfield{author}{\bibinfo{person}{Xishuang Dong}, \bibinfo{person}{Uboho
  Victor}, \bibinfo{person}{Shanta Chowdhury}, {and} \bibinfo{person}{Lijun
  Qian}.} \bibinfo{year}{2019}\natexlab{}.
\newblock \showarticletitle{Deep Two-path Semi-supervised Learning for Fake
  News Detection}.
\newblock \bibinfo{journal}{\emph{arXiv preprint arXiv:1906.05659}}
  (\bibinfo{year}{2019}).
\newblock


\bibitem[\protect\citeauthoryear{Elsken, Metzen, and Hutter}{Elsken
  et~al\mbox{.}}{2018}]%
        {elsken2018neural}
\bibfield{author}{\bibinfo{person}{Thomas Elsken}, \bibinfo{person}{Jan~Hendrik
  Metzen}, {and} \bibinfo{person}{Frank Hutter}.}
  \bibinfo{year}{2018}\natexlab{}.
\newblock \showarticletitle{Neural architecture search: A survey}.
\newblock \bibinfo{journal}{\emph{arXiv preprint arXiv:1808.05377}}
  (\bibinfo{year}{2018}).
\newblock


\bibitem[\protect\citeauthoryear{Esteban}{Esteban}{2019}]%
        {esteban2019}
\bibfield{author}{\bibinfo{person}{Ortiz-Ospina Esteban}.}
  \bibinfo{year}{2019}\natexlab{}.
\newblock \bibinfo{title}{The rise of social media}.
\newblock
\newblock
\newblock
\shownote{\url{https://ourworldindata.org/rise-of-social-media}.}


\bibitem[\protect\citeauthoryear{FactCheck}{FactCheck}{2020}]%
        {fact2020mis}
\bibfield{author}{\bibinfo{person}{FactCheck}.}
  \bibinfo{year}{2020}\natexlab{}.
\newblock \bibinfo{title}{Coronavirus Misinformation Spreads Like a Virus}.
\newblock
\newblock
\newblock
\shownote{\url{https://www.factcheck.org/2020/01/coronavirus-misinformation-spreads-like-a-virus/}.}


\bibitem[\protect\citeauthoryear{Friggeri, Adamic, Eckles, and Cheng}{Friggeri
  et~al\mbox{.}}{2014}]%
        {friggeri2014rumor}
\bibfield{author}{\bibinfo{person}{Adrien Friggeri}, \bibinfo{person}{Lada
  Adamic}, \bibinfo{person}{Dean Eckles}, {and} \bibinfo{person}{Justin
  Cheng}.} \bibinfo{year}{2014}\natexlab{}.
\newblock \showarticletitle{Rumor cascades}. In
  \bibinfo{booktitle}{\emph{Eighth International AAAI Conference on Weblogs and
  Social Media}}.
\newblock


\bibitem[\protect\citeauthoryear{Friston}{Friston}{2009}]%
        {friston2009free}
\bibfield{author}{\bibinfo{person}{Karl Friston}.}
  \bibinfo{year}{2009}\natexlab{}.
\newblock \showarticletitle{The free-energy principle: a rough guide to the
  brain?}
\newblock \bibinfo{journal}{\emph{Trends in cognitive sciences}}
  \bibinfo{volume}{13}, \bibinfo{number}{7} (\bibinfo{year}{2009}),
  \bibinfo{pages}{293--301}.
\newblock


\bibitem[\protect\citeauthoryear{Gorrell, Bontcheva, Derczynski, Kochkina,
  Liakata, and Zubiaga}{Gorrell et~al\mbox{.}}{2018}]%
        {gorrell2018rumoureval}
\bibfield{author}{\bibinfo{person}{Genevieve Gorrell}, \bibinfo{person}{Kalina
  Bontcheva}, \bibinfo{person}{Leon Derczynski}, \bibinfo{person}{Elena
  Kochkina}, \bibinfo{person}{Maria Liakata}, {and} \bibinfo{person}{Arkaitz
  Zubiaga}.} \bibinfo{year}{2018}\natexlab{}.
\newblock \showarticletitle{RumourEval 2019: Determining Rumour Veracity and
  Support for Rumours}.
\newblock \bibinfo{journal}{\emph{arXiv preprint arXiv:1809.06683}}
  (\bibinfo{year}{2018}).
\newblock


\bibitem[\protect\citeauthoryear{Gupta, Zhao, and Han}{Gupta
  et~al\mbox{.}}{2012}]%
        {gupta2012evaluating}
\bibfield{author}{\bibinfo{person}{Manish Gupta}, \bibinfo{person}{Peixiang
  Zhao}, {and} \bibinfo{person}{Jiawei Han}.} \bibinfo{year}{2012}\natexlab{}.
\newblock \showarticletitle{Evaluating event credibility on twitter}. In
  \bibinfo{booktitle}{\emph{Proceedings of the 2012 SIAM International
  Conference on Data Mining}}. SIAM, \bibinfo{pages}{153--164}.
\newblock


\bibitem[\protect\citeauthoryear{Hamidian and Diab}{Hamidian and Diab}{2015}]%
        {hamidian2015rumor}
\bibfield{author}{\bibinfo{person}{Sardar Hamidian} {and}
  \bibinfo{person}{Mona~T Diab}.} \bibinfo{year}{2015}\natexlab{}.
\newblock \showarticletitle{Rumor detection and classification for twitter
  data}. In \bibinfo{booktitle}{\emph{Proceedings of the Fifth International
  Conference on Social Media Technologies, Communication, and Informatics
  (SOTICS)}}. \bibinfo{pages}{71--77}.
\newblock


\bibitem[\protect\citeauthoryear{Han, Gao, and Ciravegna}{Han
  et~al\mbox{.}}{2019}]%
        {han2019data}
\bibfield{author}{\bibinfo{person}{Sooji Han}, \bibinfo{person}{Jie Gao}, {and}
  \bibinfo{person}{Fabio Ciravegna}.} \bibinfo{year}{2019}\natexlab{}.
\newblock \showarticletitle{Data Augmentation for Rumor Detection Using
  Context-Sensitive Neural Language Model With Large-Scale Credibility Corpus}.
\newblock  (\bibinfo{year}{2019}).
\newblock


\bibitem[\protect\citeauthoryear{Hochreiter and Schmidhuber}{Hochreiter and
  Schmidhuber}{1997}]%
        {hochreiter1997long}
\bibfield{author}{\bibinfo{person}{Sepp Hochreiter} {and}
  \bibinfo{person}{J{\"u}rgen Schmidhuber}.} \bibinfo{year}{1997}\natexlab{}.
\newblock \showarticletitle{Long short-term memory}.
\newblock \bibinfo{journal}{\emph{Neural computation}} \bibinfo{volume}{9},
  \bibinfo{number}{8} (\bibinfo{year}{1997}), \bibinfo{pages}{1735--1780}.
\newblock


\bibitem[\protect\citeauthoryear{Jenn}{Jenn}{2009}]%
        {Jenn2009}
\bibfield{author}{\bibinfo{person}{Lowther Jenn}.}
  \bibinfo{year}{2009}\natexlab{}.
\newblock \bibinfo{title}{Microblogging is one of the top four trends in social
  media}.
\newblock
\newblock
\newblock
\shownote{\url{https://www.straight.com/article-200494/microblogging-one-top-four-trends-social-media}.}


\bibitem[\protect\citeauthoryear{Jessica}{Jessica}{2020}]%
        {jessica2020wuhan}
\bibfield{author}{\bibinfo{person}{McDonald Jessica}.}
  \bibinfo{year}{2020}\natexlab{}.
\newblock \bibinfo{title}{Q\&A on the Wuhan Coronavirus}.
\newblock
\newblock
\newblock
\shownote{\url{https://www.factcheck.org/2020/01/qa-on-the-wuhan-coronavirus/}.}


\bibitem[\protect\citeauthoryear{Jin, Cao, Guo, Zhang, Wang, and Luo}{Jin
  et~al\mbox{.}}{2017}]%
        {jin2017detection}
\bibfield{author}{\bibinfo{person}{Zhiwei Jin}, \bibinfo{person}{Juan Cao},
  \bibinfo{person}{Han Guo}, \bibinfo{person}{Yongdong Zhang},
  \bibinfo{person}{Yu Wang}, {and} \bibinfo{person}{Jiebo Luo}.}
  \bibinfo{year}{2017}\natexlab{}.
\newblock \showarticletitle{Detection and analysis of 2016 us presidential
  election related rumors on twitter}. In
  \bibinfo{booktitle}{\emph{International conference on social computing,
  behavioral-cultural modeling and prediction and behavior representation in
  modeling and simulation}}. Springer, \bibinfo{pages}{14--24}.
\newblock


\bibitem[\protect\citeauthoryear{Julie}{Julie}{2020}]%
        {Julie2020}
\bibfield{author}{\bibinfo{person}{Wernau Julie}.}
  \bibinfo{year}{2020}\natexlab{}.
\newblock \bibinfo{title}{Virus Sparks Chinese Panic Buying, Travel
  Cancellations and Social-Media Misinformation}.
\newblock
\newblock
\newblock
\shownote{\url{https://www.wsj.com/articles/coronavirus-sparks-chinese-panic-buying-travel-cancellations-and-social-media-misinformation-11579698948}.}


\bibitem[\protect\citeauthoryear{Katerina and Elisa}{Katerina and
  Elisa}{2018}]%
        {pewresearch}
\bibfield{author}{\bibinfo{person}{Matsa Katerina} {and}
  \bibinfo{person}{Shearer Elisa}.} \bibinfo{year}{2018}\natexlab{}.
\newblock \bibinfo{title}{News Use Across Social Media Platforms 2018}.
\newblock
\newblock
\urldef\tempurl%
\url{https://www.journalism.org/2018/09/10/news-use-across-social-media-platforms-2018/}
\showURL{%
Retrieved September 9, 2019 from \tempurl}


\bibitem[\protect\citeauthoryear{Kingma and Welling}{Kingma and
  Welling}{2013}]%
        {kingma2013auto}
\bibfield{author}{\bibinfo{person}{Diederik~P Kingma} {and}
  \bibinfo{person}{Max Welling}.} \bibinfo{year}{2013}\natexlab{}.
\newblock \showarticletitle{Auto-encoding variational bayes}.
\newblock \bibinfo{journal}{\emph{arXiv preprint arXiv:1312.6114}}
  (\bibinfo{year}{2013}).
\newblock


\bibitem[\protect\citeauthoryear{Kochkina, Liakata, and Zubiaga}{Kochkina
  et~al\mbox{.}}{2018}]%
        {kochkina2018all}
\bibfield{author}{\bibinfo{person}{Elena Kochkina}, \bibinfo{person}{Maria
  Liakata}, {and} \bibinfo{person}{Arkaitz Zubiaga}.}
  \bibinfo{year}{2018}\natexlab{}.
\newblock \showarticletitle{All-in-one: Multi-task learning for rumour
  verification}.
\newblock \bibinfo{journal}{\emph{arXiv preprint arXiv:1806.03713}}
  (\bibinfo{year}{2018}).
\newblock


\bibitem[\protect\citeauthoryear{Kumar and Carley}{Kumar and Carley}{2019}]%
        {kumar2019tree}
\bibfield{author}{\bibinfo{person}{Sumeet Kumar} {and}
  \bibinfo{person}{Kathleen~M Carley}.} \bibinfo{year}{2019}\natexlab{}.
\newblock \showarticletitle{Tree LSTMs with Convolution Units to Predict Stance
  and Rumor Veracity in Social Media Conversations}. In
  \bibinfo{booktitle}{\emph{Proceedings of the 57th Conference of the
  Association for Computational Linguistics}}. \bibinfo{pages}{5047--5058}.
\newblock


\bibitem[\protect\citeauthoryear{Kwon, Cha, Jung, Chen, and Wang}{Kwon
  et~al\mbox{.}}{2013}]%
        {kwon2013prominent}
\bibfield{author}{\bibinfo{person}{Sejeong Kwon}, \bibinfo{person}{Meeyoung
  Cha}, \bibinfo{person}{Kyomin Jung}, \bibinfo{person}{Wei Chen}, {and}
  \bibinfo{person}{Yajun Wang}.} \bibinfo{year}{2013}\natexlab{}.
\newblock \showarticletitle{Prominent features of rumor propagation in online
  social media}. In \bibinfo{booktitle}{\emph{2013 IEEE 13th International
  Conference on Data Mining}}. IEEE, \bibinfo{pages}{1103--1108}.
\newblock


\bibitem[\protect\citeauthoryear{Lewandowsky, Ecker, Seifert, Schwarz, and
  Cook}{Lewandowsky et~al\mbox{.}}{2012}]%
        {lewandowsky2012misinformation}
\bibfield{author}{\bibinfo{person}{Stephan Lewandowsky},
  \bibinfo{person}{Ullrich~KH Ecker}, \bibinfo{person}{Colleen~M Seifert},
  \bibinfo{person}{Norbert Schwarz}, {and} \bibinfo{person}{John Cook}.}
  \bibinfo{year}{2012}\natexlab{}.
\newblock \showarticletitle{Misinformation and its correction: Continued
  influence and successful debiasing}.
\newblock \bibinfo{journal}{\emph{Psychological Science in the Public
  Interest}} \bibinfo{volume}{13}, \bibinfo{number}{3} (\bibinfo{year}{2012}),
  \bibinfo{pages}{106--131}.
\newblock


\bibitem[\protect\citeauthoryear{Li, Zhang, and Si}{Li et~al\mbox{.}}{2019}]%
        {li2019rumor}
\bibfield{author}{\bibinfo{person}{Quanzhi Li}, \bibinfo{person}{Qiong Zhang},
  {and} \bibinfo{person}{Luo Si}.} \bibinfo{year}{2019}\natexlab{}.
\newblock \showarticletitle{Rumor Detection by Exploiting User Credibility
  Information, Attention and Multi-task Learning}. In
  \bibinfo{booktitle}{\emph{Proceedings of the 57th Annual Meeting of the
  Association for Computational Linguistics}}. \bibinfo{pages}{1173--1179}.
\newblock


\bibitem[\protect\citeauthoryear{Li, Sun, and Zhu}{Li et~al\mbox{.}}{2010}]%
        {li2010data}
\bibfield{author}{\bibinfo{person}{Yanling Li}, \bibinfo{person}{Guoshe Sun},
  {and} \bibinfo{person}{Yehang Zhu}.} \bibinfo{year}{2010}\natexlab{}.
\newblock \showarticletitle{Data imbalance problem in text classification}. In
  \bibinfo{booktitle}{\emph{2010 Third International Symposium on Information
  Processing}}. IEEE, \bibinfo{pages}{301--305}.
\newblock


\bibitem[\protect\citeauthoryear{Liu, Simonyan, Vinyals, Fernando, and
  Kavukcuoglu}{Liu et~al\mbox{.}}{2017}]%
        {liu2017hierarchical}
\bibfield{author}{\bibinfo{person}{Hanxiao Liu}, \bibinfo{person}{Karen
  Simonyan}, \bibinfo{person}{Oriol Vinyals}, \bibinfo{person}{Chrisantha
  Fernando}, {and} \bibinfo{person}{Koray Kavukcuoglu}.}
  \bibinfo{year}{2017}\natexlab{}.
\newblock \showarticletitle{Hierarchical representations for efficient
  architecture search}.
\newblock \bibinfo{journal}{\emph{arXiv preprint arXiv:1711.00436}}
  (\bibinfo{year}{2017}).
\newblock


\bibitem[\protect\citeauthoryear{Ma, Gao, and Wong}{Ma et~al\mbox{.}}{2018}]%
        {ma2018detect}
\bibfield{author}{\bibinfo{person}{Jing Ma}, \bibinfo{person}{Wei Gao}, {and}
  \bibinfo{person}{Kam-Fai Wong}.} \bibinfo{year}{2018}\natexlab{}.
\newblock \showarticletitle{Detect rumor and stance jointly by neural
  multi-task learning}. In \bibinfo{booktitle}{\emph{Companion Proceedings of
  the The Web Conference 2018}}. International World Wide Web Conferences
  Steering Committee, \bibinfo{pages}{585--593}.
\newblock


\bibitem[\protect\citeauthoryear{Ma, Gao, and Wong}{Ma et~al\mbox{.}}{2019}]%
        {ma2019detect}
\bibfield{author}{\bibinfo{person}{Jing Ma}, \bibinfo{person}{Wei Gao}, {and}
  \bibinfo{person}{Kam-Fai Wong}.} \bibinfo{year}{2019}\natexlab{}.
\newblock \showarticletitle{Detect Rumors on Twitter by Promoting Information
  Campaigns with Generative Adversarial Learning}. In
  \bibinfo{booktitle}{\emph{The World Wide Web Conference}}. ACM,
  \bibinfo{pages}{3049--3055}.
\newblock


\bibitem[\protect\citeauthoryear{Matt}{Matt}{2020}]%
        {matt2020fake}
\bibfield{author}{\bibinfo{person}{Field Matt}.}
  \bibinfo{year}{2020}\natexlab{}.
\newblock \bibinfo{title}{Fake news epidemic: Coronavirus breeds hate and
  disinformation in India and beyond}.
\newblock
\newblock
\newblock
\shownote{\url{https://thebulletin.org/2020/01/fake-news-epidemic-coronavirus-breeds-hate-and-disinformation-in-india-and-beyond/}.}


\bibitem[\protect\citeauthoryear{Mercey}{Mercey}{2020}]%
        {mercey2020corona}
\bibfield{author}{\bibinfo{person}{Livingston Mercey}.}
  \bibinfo{year}{2020}\natexlab{}.
\newblock \bibinfo{title}{Coronavirus fact check: How to spot fake reports
  about the mysterious disease}.
\newblock
\newblock
\newblock
\shownote{\url{https://www.cnet.com/how-to/false-information-about-coronavirus-here-are-the-top-rumors-spreading-about-it/}.}


\bibitem[\protect\citeauthoryear{Nguyen, Do, Calderbank, and
  Deligiannis}{Nguyen et~al\mbox{.}}{2019}]%
        {nguyen2019fake}
\bibfield{author}{\bibinfo{person}{Duc~Minh Nguyen}, \bibinfo{person}{Tien~Huu
  Do}, \bibinfo{person}{Robert Calderbank}, {and} \bibinfo{person}{Nikos
  Deligiannis}.} \bibinfo{year}{2019}\natexlab{}.
\newblock \showarticletitle{Fake News Detection using Deep Markov Random
  Fields}. In \bibinfo{booktitle}{\emph{Proceedings of the 2019 Conference of
  the North American Chapter of the Association for Computational Linguistics:
  Human Language Technologies, Volume 1 (Long and Short Papers)}}.
  \bibinfo{pages}{1391--1400}.
\newblock


\bibitem[\protect\citeauthoryear{Oshikawa, Qian, and Wang}{Oshikawa
  et~al\mbox{.}}{2018}]%
        {oshikawa2018survey}
\bibfield{author}{\bibinfo{person}{Ray Oshikawa}, \bibinfo{person}{Jing Qian},
  {and} \bibinfo{person}{William~Yang Wang}.} \bibinfo{year}{2018}\natexlab{}.
\newblock \showarticletitle{A survey on natural language processing for fake
  news detection}.
\newblock \bibinfo{journal}{\emph{arXiv preprint arXiv:1811.00770}}
  (\bibinfo{year}{2018}).
\newblock


\bibitem[\protect\citeauthoryear{Real, Aggarwal, Huang, and Le}{Real
  et~al\mbox{.}}{2019}]%
        {real2019regularized}
\bibfield{author}{\bibinfo{person}{Esteban Real}, \bibinfo{person}{Alok
  Aggarwal}, \bibinfo{person}{Yanping Huang}, {and} \bibinfo{person}{Quoc~V
  Le}.} \bibinfo{year}{2019}\natexlab{}.
\newblock \showarticletitle{Regularized evolution for image classifier
  architecture search}. In \bibinfo{booktitle}{\emph{Proceedings of the AAAI
  Conference on Artificial Intelligence}}, Vol.~\bibinfo{volume}{33}.
  \bibinfo{pages}{4780--4789}.
\newblock


\bibitem[\protect\citeauthoryear{Sadiq, Wagner, Shyu, and Feaster}{Sadiq
  et~al\mbox{.}}{2019}]%
        {sadiq2019high}
\bibfield{author}{\bibinfo{person}{Saad Sadiq}, \bibinfo{person}{Nicolas
  Wagner}, \bibinfo{person}{Mei-Ling Shyu}, {and} \bibinfo{person}{Daniel
  Feaster}.} \bibinfo{year}{2019}\natexlab{}.
\newblock \showarticletitle{High Dimensional Latent Space Variational
  AutoEncoders for Fake News Detection}. In \bibinfo{booktitle}{\emph{2019 IEEE
  Conference on Multimedia Information Processing and Retrieval (MIPR)}}. IEEE,
  \bibinfo{pages}{437--442}.
\newblock


\bibitem[\protect\citeauthoryear{Tony}{Tony}{2020}]%
        {tony2020mis}
\bibfield{author}{\bibinfo{person}{Romm Tony}.}
  \bibinfo{year}{2020}\natexlab{}.
\newblock \bibinfo{title}{Facebook will remove misinformation about
  coronavirus}.
\newblock
\newblock
\newblock
\shownote{\url{https://www.washingtonpost.com/technology/2020/01/30/facebook-coronavirus-fakes/}.}


\bibitem[\protect\citeauthoryear{Van~Asch}{Van~Asch}{2013}]%
        {van2013macro}
\bibfield{author}{\bibinfo{person}{Vincent Van~Asch}.}
  \bibinfo{year}{2013}\natexlab{}.
\newblock \showarticletitle{Macro-and micro-averaged evaluation measures
  [[basic draft]]}.
\newblock \bibinfo{journal}{\emph{Belgium: CLiPS}}  \bibinfo{volume}{49}
  (\bibinfo{year}{2013}).
\newblock


\bibitem[\protect\citeauthoryear{Zhang, Lipani, Liang, and Yilmaz}{Zhang
  et~al\mbox{.}}{2019}]%
        {zhang2019reply}
\bibfield{author}{\bibinfo{person}{Qiang Zhang}, \bibinfo{person}{Aldo Lipani},
  \bibinfo{person}{Shangsong Liang}, {and} \bibinfo{person}{Emine Yilmaz}.}
  \bibinfo{year}{2019}\natexlab{}.
\newblock \showarticletitle{Reply-Aided Detection of Misinformation via
  Bayesian Deep Learning}. In \bibinfo{booktitle}{\emph{The World Wide Web
  Conference}}. ACM, \bibinfo{pages}{2333--2343}.
\newblock


\bibitem[\protect\citeauthoryear{Zoe}{Zoe}{2020}]%
        {zoe2020fb}
\bibfield{author}{\bibinfo{person}{Thomas Zoe}.}
  \bibinfo{year}{2020}\natexlab{}.
\newblock \bibinfo{title}{Coronavirus: How Facebook, TikTok and other apps
  tackle fake claims}.
\newblock
\newblock
\newblock
\shownote{\url{https://www.bbc.com/news/technology-51337357}.}


\bibitem[\protect\citeauthoryear{Zubiaga, Aker, Bontcheva, Liakata, and
  Procter}{Zubiaga et~al\mbox{.}}{2018}]%
        {zubiaga2018detection}
\bibfield{author}{\bibinfo{person}{Arkaitz Zubiaga}, \bibinfo{person}{Ahmet
  Aker}, \bibinfo{person}{Kalina Bontcheva}, \bibinfo{person}{Maria Liakata},
  {and} \bibinfo{person}{Rob Procter}.} \bibinfo{year}{2018}\natexlab{}.
\newblock \showarticletitle{Detection and resolution of rumours in social
  media: A survey}.
\newblock \bibinfo{journal}{\emph{ACM Computing Surveys (CSUR)}}
  \bibinfo{volume}{51}, \bibinfo{number}{2} (\bibinfo{year}{2018}),
  \bibinfo{pages}{32}.
\newblock


\bibitem[\protect\citeauthoryear{Zubiaga, Kochkina, Liakata, Procter, and
  Lukasik}{Zubiaga et~al\mbox{.}}{2016}]%
        {zubiaga2016stance}
\bibfield{author}{\bibinfo{person}{Arkaitz Zubiaga}, \bibinfo{person}{Elena
  Kochkina}, \bibinfo{person}{Maria Liakata}, \bibinfo{person}{Rob Procter},
  {and} \bibinfo{person}{Michal Lukasik}.} \bibinfo{year}{2016}\natexlab{}.
\newblock \showarticletitle{Stance classification in rumours as a sequential
  task exploiting the tree structure of social media conversations}.
\newblock \bibinfo{journal}{\emph{arXiv preprint arXiv:1609.09028}}
  (\bibinfo{year}{2016}).
\newblock


\bibitem[\protect\citeauthoryear{Zubiaga, Liakata, and Procter}{Zubiaga
  et~al\mbox{.}}{2017}]%
        {zubiaga2017exploiting}
\bibfield{author}{\bibinfo{person}{Arkaitz Zubiaga}, \bibinfo{person}{Maria
  Liakata}, {and} \bibinfo{person}{Rob Procter}.}
  \bibinfo{year}{2017}\natexlab{}.
\newblock \showarticletitle{Exploiting context for rumour detection in social
  media}. In \bibinfo{booktitle}{\emph{International Conference on Social
  Informatics}}. Springer, \bibinfo{pages}{109--123}.
\newblock


\end{thebibliography}


\end{document}